\documentclass[lettersize,journal]{IEEEtran}
\usepackage{amsmath,amsfonts}
\usepackage{algorithmic}
\usepackage{algorithm}
\usepackage{array}
\usepackage[caption=false,font=normalsize,labelfont=sf,textfont=sf]{subfig}
\usepackage{textcomp}
\usepackage{stfloats}
\usepackage{url}
\usepackage{verbatim}
\usepackage{graphicx}
\usepackage{cite}
\usepackage{booktabs}
\usepackage{color}    

\begin{document}

\title{FS-DVS: A Frequency-Selective Dynamic Visual Sensing Paradigm for Enhancing Information Completeness}

\author{
    \IEEEauthorblockN{
        Feiyu Ji,
        Xiaokang Yang,~\IEEEmembership{Fellow,~IEEE,}
        Xiaoyun Yuan*,~\IEEEmembership{Member,~IEEE,}\thanks{*Corresponding author.}
    }\\
    \IEEEauthorblockA{
        MoE Key Lab of Artificial Intelligence, Institute of AI, School of Computer Science, Shanghai Jiao Tong University\\
        Email: \{jownr25, xkyang, yuanxiaoyun\}@sjtu.edu.cn
    }
}


\markboth{Journal of \LaTeX\ Class Files,~Vol.~14, No.~8, August~2021}%
{Shell \MakeLowercase{\textit{et al.}}: A Sample Article Using IEEEtran.cls for IEEE Journals}


\maketitle

\begin{abstract}
Dynamic vision sensors (DVS) offer exceptional temporal resolution and dynamic range by asynchronously reporting pixel-level intensity changes. However, conventional DVS rely on a per-pixel independent triggering mechanism, ignoring the spatial integration performed by biological retinal ganglion cells (RGCs). Consequently, they lack the contrast sensitivity function (CSF) and its inherent sensitivity to mid-spatial frequencies, which inevitably leads to information incompleteness due to sub-threshold signal loss. To bridge this gap, we propose FS-DVS (Frequency-Selective Dynamic Vision Sensor), a novel paradigm that integrates a learnable spatial filter strictly preceding the event triggering process to mimic the RGC aggregation mechanism. By developing a differentiable event simulation framework, the spatial filter can be optimized end-to-end with downstream tasks. Our study reveals that starting from a delta function, the learned spatial filters spontaneously evolve into center-surround patterns that emphasize mid-frequency components, consistently aligning with human CSF. Beyond achieving substantial performance gains in object detection and action recognition, the consistent convergence to human-like CSF characteristics across different tasks underscores the universality of this mid-frequency selective mechanism. Compared to naively increasing sensor sensitivity or relying on post-processing, our paradigm achieves selective information enhancement with high noise resilience, providing a robust, biologically plausible blueprint for next-generation neuromorphic sensors.

\end{abstract}

\begin{IEEEkeywords}
Dynamic vision sensors, neuromorphic vision, frequency selectivity, contrast sensitivity function
\end{IEEEkeywords}

\section{Introduction}
\IEEEPARstart{D}{ynamic} vision sensors (DVS) are neuromorphic vision sensors that differ fundamentally from conventional frame-based cameras. Instead of sampling intensity frames at fixed rates, they asynchronously emit events in response to logarithmic intensity changes at individual pixels~\cite{lichtsteiner2008128}. This event-driven sensing paradigm offers microsecond-level temporal resolution, high dynamic range, and low latency~\cite{gallego2020event}. However, conventional DVS rely on an independent per-pixel triggering mechanism, which ignores the spatial integration inherent in the biological retina~\cite{wandell1995foundations}. In the human visual system, retinal ganglion cells (RGCs) aggregate signals from multiple photoreceptors within local receptive fields before generating spikes~\cite{kuffler1953discharge}. This spatial integration yields the contrast sensitivity function (CSF), characterized by a pronounced sensitivity to mid-spatial frequencies (representing physical topologies and contours) and a natural suppression of low-frequency backgrounds and high-frequency noise~\cite{campbell1968application}.

The absence of such a frequency-selective mechanism in conventional DVS leads to a fundamental bottleneck: structural incompleteness. Due to the independent pixel-wise triggering, weak structural edges often fail to accumulate sufficient intensity variations to cross the rigid contrast threshold, yielding highly fragmented spatio-temporal representations. Crucially, this limitation cannot be resolved by simply increasing global hardware sensitivity, which triggers a noise avalanche due to the spatio-temporal ambiguity of single-pixel sensing~\cite{delbruck2021feedback}, nor by post-processing, which cannot retrieve structural information permanently truncated at the sensing source.

To address these limitations, we propose FS-DVS (Frequency-Selective Dynamic Vision Sensor), a novel sensing paradigm that integrates a learnable spatial filter prior to the event triggering mechanism to mimic the spatial aggregation of RGCs. In addition, we develop a fully differentiable event simulation framework, allowing the spatial filter parameters to be jointly optimized with downstream models, transforming event generation from a fixed detection process into a task-adaptive perceptual encoding mechanism.

A particularly interesting finding of this work is that when initialized with a delta-like point-spread function and driven purely by downstream tasks such as detection and recognition, the learned spatial filters spontaneously converge to a center-enhanced, surround-suppressed morphology. In the frequency domain, this evolution exhibits a mid-frequency emphasis that closely aligns with human CSF, despite the absence of any explicit biological constraints during training. Beyond achieving substantial performance gains in object detection and action recognition, the learned filters' consistent convergence to human-like characteristics across tasks underscores the universality of this mid-frequency selective mechanism. This phenomenon suggests that frequency-selective perception is not merely a biological coincidence but a mathematically optimal strategy for maximizing task-driven information acquisition while suppressing noise.

We systematically validate the FS-DVS paradigm through a multi-dimensional experimental storyline. We first evaluate FS-DVS on object detection and action recognition, where it consistently improves performance over conventional DVS and drives the learned filters toward human-like CSF characteristics across different tasks. This consistent task-driven convergence motivates us to further examine the universality of the learned mid-frequency selective mechanism. Experiments on semantic segmentation show that filters learned from detection tasks generalize effectively to dense perception tasks, suggesting that the learned mid-frequency selectivity is not confined to a specific optimization objective but reflects a transferable sensing mechanism. Finally, we conduct mutual information (MI) analysis as an information-theoretic validation, demonstrating that FS-DVS preserves more effective information than conventional DVS. Together, these results establish FS-DVS as a robust and biologically plausible sensing paradigm for next-generation neuromorphic vision sensors.

In summary, the main contributions of this work are as follows:
\begin{itemize}
\item We propose the FS-DVS paradigm along with a fully differentiable event simulation framework, enabling the end-to-end optimization of a spatial filter that mimics the RGC aggregation mechanism prior to event triggering.
\item We demonstrate that FS-DVS achieves substantial performance gains (e.g., 12.3\% mAP in detection) by restoring structural completeness, outperforming conventional DVS in noise-robust structural preservation and effective information acquisition.

\item We discover that task-driven optimization leads to the spontaneous convergence of spatial filters to a mid-frequency emphasis consistent with human CSF, and verify the universality of this mechanism across diverse tasks including detection, recognition, and segmentation.
\end{itemize}

\section{Related Work}
\subsection{Dynamic Vision Sensors}
The silicon retina concept, inspired by the asynchronous nature of human vision, laid the foundation for dynamic vision sensors (DVS)~\cite{mahowald1992vlsi}. Unlike conventional frame-based cameras, DVS pixels independently monitor relative intensity changes and emit asynchronous event streams when these changes exceed a predefined threshold~\cite{posch2014retinomorphic, delbruck2016neuromorophic}. To enrich the sparse event data, architectural variants such as ATIS~\cite{posch2010qvga} and DAVIS~\cite{brandli2014240} were developed to incorporate absolute intensity or intensity-change information. Crucially, despite these architectural additions, all these conventional sensors still adhere to a strictly per-pixel sensing paradigm, where each pixel operates independently without lateral communication. 

In contrast, biological vision systems rely on retinal ganglion cells (RGCs) that integrate information across a spatial receptive field to determine whether to trigger a spike. Inspired by this biological paradigm, recent works have explored incorporating spatial context. 
For instance, center-surround DVS (CSDVS)~\cite{delbruck2022utility} proposes a hardware design using lateral resistance to suppress low-frequency redundancy and amplify high-frequency transients, which was verified in simulation.
The neural ganglion sensor (NGS)~\cite{so2025neural} proposed a learning-based algorithm to optimize the spatial kernel. Yet, NGS was primarily validated on specific low-level tasks (e.g., video interpolation and optical flow) and remained confined to simulation.

Moving beyond these approaches, our FS-DVS introduces a physical sensor-in-the-loop verification. We demonstrate the strong cross-task generalizability of the learned spatial filters across diverse visual applications. Notably, our analysis reveals that these optimized filters universally converge to a mid-frequency enhancement mechanism, aligning with the contrast sensitivity of human visual perception.

\subsection{Dynamic Vision Sensor Simulation and Modeling}
The widespread availability of standard cameras makes conventional frame-based data exceptionally abundant. In stark contrast, event data acquisition is heavily bottlenecked by the need for specialized dynamic vision sensors. This hardware dependency, compounded by the difficulty of annotating sparse event signals, makes large-scale event datasets exceedingly rare. To resolve this, simulators have been developed to convert video sequences into event streams via rule-based (e.g., ESIM~\cite{rebecq2018esim}, V2E~\cite{hu2021v2e}) or learning-based (e.g., EventGAN~\cite{zhu2021eventgan}, V2CE~\cite{zhang2024v2ce}) approaches. While rule-based methods rely on logarithmic intensity models with idealized sensor assumptions, learning-based methods can improve accuracy and realism by implicitly modeling non-ideal characteristics of real-world sensors from data. Despite these advances, most current simulators function merely as standalone data generation pipelines. There remains a critical need for highly realistic, fully differentiable event modeling frameworks that can be jointly optimized with downstream neural networks.

\subsection{Event Representation and Processing}

Dynamic vision sensors output asynchronous events in the form of tuples $e = (x, y, t, p)$, where $t$ denotes the timestamp, $(x, y)$ represent the spatial coordinates, and $p$ indicates the polarity of the intensity change. Directly operating on such raw event streams often requires specialized processing architectures, among which spiking neural networks (SNNs) provide a naturally event-driven paradigm. While SNNs have proven effective for various vision tasks~\cite{zhou2022spikformer, su2023deep, zhu2022event}, their optimization remains exceptionally challenging due to the inherent non-differentiability of discrete spikes, and their performance generally lags behind that of conventional artificial neural networks (ANNs).
Consequently, event streams are commonly converted into dense, tensor-based representations that are compatible with established deep learning backbones. The most straightforward approach is to accumulate events within a fixed temporal window into an event frame of shape $2 \times H \times W$, allocating one channel for each polarity~\cite{perot2020learning}. Another widely adopted representation is the event voxel grid, where events are discretized into a three-dimensional spatio-temporal tensor of shape $B \times H \times W$, with $B$ denoting the number of temporal bins~\cite{zhu2019unsupervised}. More recently, optimized dense representations have been explored to better preserve the structural fidelity of raw streams while maintaining compatibility with standard vision models~\cite{zubic2023chaos, 10758409}. Ultimately, these representations offer varying trade-offs among temporal precision, downstream task performance, and computational overhead.

Building upon these representation paradigms, downstream event processing tasks have developed along both the event-native and tensor-based tracks. For object detection, specialized frameworks utilize either SNN-based structures~\cite{kim2020spiking, luo2024integer} or ANN-based architectures featuring recurrent memories and Transformers, such as RVT~\cite{gehrig2023recurrent}, GET~\cite{peng2023get}, and SAST~\cite{peng2024scene}, to capture spatio-temporal context. More recently, adaptation frameworks like I2EvDet~\cite{torbunov2025evrt} have demonstrated that applying mature image-based detectors to standard dense event frames can achieve performance on par with dedicated event-native detectors.

Similarly, event-based action recognition spans both parallel computing tracks. While event-native networks like graph convolutional networks (GCNs)~\cite{9672728} exploit raw spatial sparsity, a dominant strategy transforms events into dense frames or voxel grids for processing via conventional CNN- or Transformer-based video architectures~\cite{gao2023action, 9857382, de2023eventtransact}. Furthermore, multimodal methods combine these event representations with RGB frames~\cite{11000281} or optical flow~\cite{9879783} to enrich appearance and motion cues for improved recognition.

Beyond detection and recognition, event-based semantic segmentation extends event processing to dense pixel-level understanding. Existing methods either construct dense event representations for pixel-level prediction~\cite{sun2022ess, jia2023event}, or maintain event-native temporal states for low-latency dense inference~\cite{hamaguchi2023hierarchical, 10058930}. These studies make segmentation a stringent benchmark for evaluating whether event representations preserve fine-grained spatial structures, object boundaries, and semantic cues.

Overall, the vast majority of existing representation and processing methods focus on optimizing downstream network architectures and data representation. In contrast, our work shifts the paradigm upstream: we investigate the event generation process itself, aiming to fundamentally enrich the structural information of the raw event streams and thereby facilitate a universal performance boost across diverse downstream perception tasks.

\begin{figure*}[h]
    \centering
    \includegraphics[width=\textwidth]{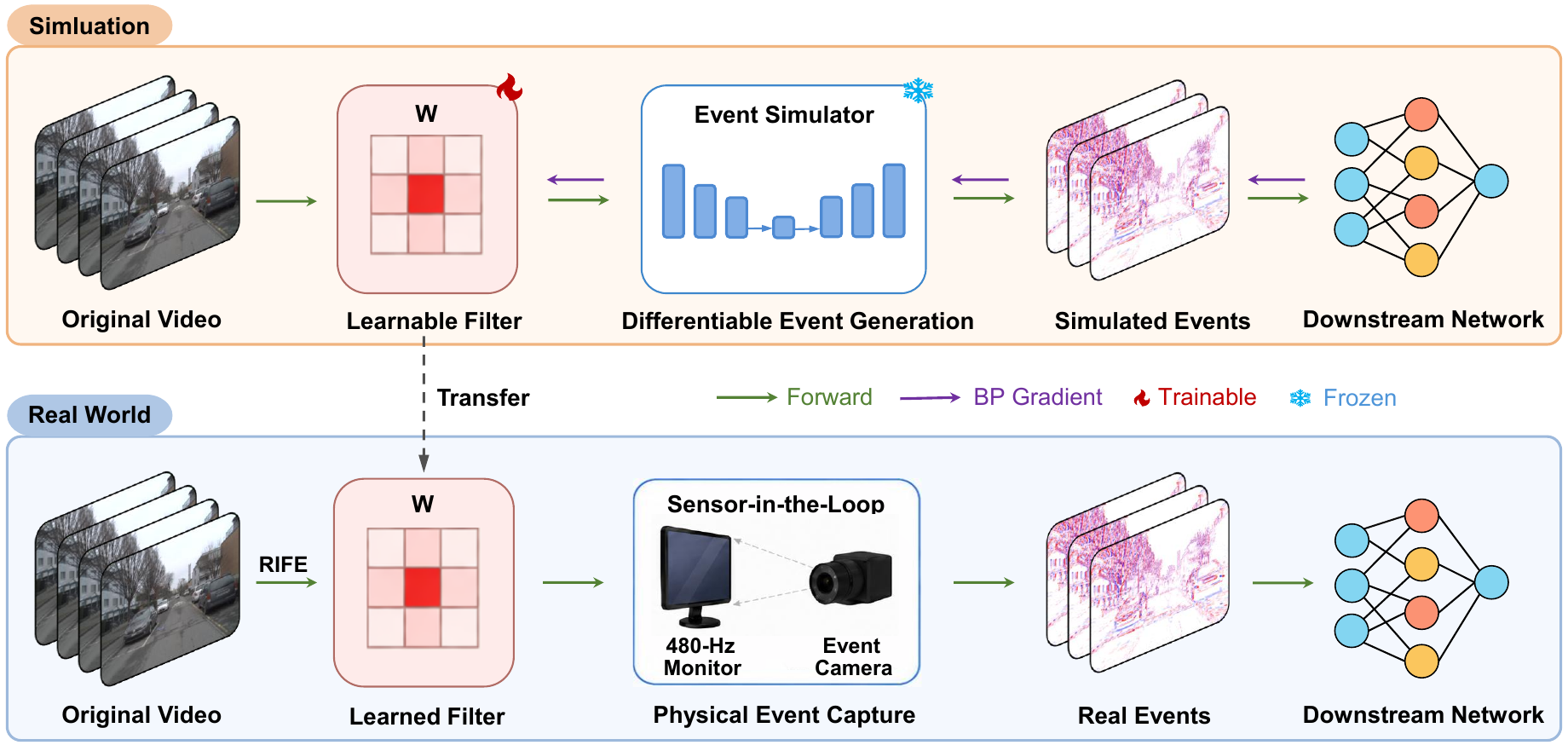}
    \caption{Overview of the proposed FS-DVS pipeline. In the simulation branch, the learnable filter is placed before event triggering and optimized through a frozen differentiable event simulator with downstream supervision. In the physical-capture branch, the original video is first temporally upsampled by RIFE~\cite{huang2022realtimeintermediateflowestimation} and then filtered by the learned spatial kernel before being displayed and recaptured in a sensor-in-the-loop setup for real-data evaluation.}
    \label{fig:fsdvs_pipeline}
\end{figure*}

\section{Proposed Method}

\subsection{Revisiting Conventional Dynamic Vision Sensors}
Conventional DVS operate under the independent pixel-wise sensing model. Let $I(\mathbf{x}, t)$ denote the image intensity at spatial location $\mathbf{x} = (x, y)^{\top}$ and time $t$, and let $L(\mathbf{x}, t) \doteq \log I(\mathbf{x}, t)$ be the corresponding logarithmic intensity. A pixel asynchronously generates an event $e_k = (\mathbf{x}_k, t_k, p_k)$ when the logarithmic intensity change since its last event at time $t_{\text{last}}$ exceeds a predefined contrast threshold $C > 0$:
\begin{equation}
    \Delta L(\mathbf{x}_k, t_k) \doteq L(\mathbf{x}_k, t_k) - L(\mathbf{x}_k, t_{\text{last}}) = p_k C,
    \label{eq:standard_dvs}
\end{equation}
where $p_k \in \{+1, -1\}$ denotes the polarity of the brightness change.
For a sufficiently small time interval $\Delta t = t_k - t_{\text{last}}$, this temporal variation can be approximated using the first-order Taylor expansion:
\begin{equation}
    \Delta L(\mathbf{x}_k, t_k) \approx \nabla L(\mathbf{x}_k, t_k) \cdot \mathbf{v}(\mathbf{x}_k, t_k) \Delta t,
    \label{eq:taylor_expansion}
\end{equation}
where $\nabla L$ is the spatial brightness gradient (representing scene edges) and $\mathbf{v}$ is the apparent motion velocity.

This formulation reveals a fundamental limitation: in areas with weak structural edges (i.e., $\|\nabla L\|$ is small), the intensity variation often fails to reach the rigid hardware threshold $C$. As a result, these sub-threshold edges are ignored by the sensor, yielding fragmented object contours.
A straightforward approach to capture these weak edges is to increase the sensor's global sensitivity by lowering the threshold $C$. However, in real sensors, the signal is invariably corrupted by thermal noise and dark currents. Blindly lowering $C$ drastically amplifies the impact of this noise, flooding the visual representation with random noise events. Therefore, resolving structural incompleteness while maintaining noise resilience inherently requires introducing spatial correlation prior to the triggering mechanism.

\subsection{FS-DVS Sensing Paradigm}
Unlike conventional DVS, human RGCs do not generate action potentials based on the response of a single photoreceptor. Instead, they aggregate signals from multiple photoreceptors within a local receptive field. This spatial integration acts as a critical first-layer physical filter prior to spike generation. From the perspective of visual signal processing, the ``center-surround'' organization of RGCs fundamentally acts as a spatial band-pass filter. This mechanism naturally attenuates low-frequency flat backgrounds and high-frequency thermal noise, while selectively emphasizing mid-spatial frequencies that correspond to physical structures and contours. Macroscopically, this frequency-selective perception manifests as the human CSF. 

Inspired by this, we propose the FS-DVS paradigm. Instead of directly thresholding isolated pixel intensities, FS-DVS introduces a spatial filter $\mathbf{W}$ prior to the logarithmic transformation and the event triggering comparator. Let $\mathbf{W}$ denote the spatial convolution kernel representing the receptive field. The spatially aggregated intensity at location $\mathbf{x}$ and time $t$ is given by:
\begin{equation}
    I_{\text{agg}}(\mathbf{x}, t) = (\mathbf{W} \ast I)(\mathbf{x}, t),
\end{equation}
where $\ast$ denotes the 2D spatial convolution operation. 
Following this spatial aggregation, the signal undergoes the standard logarithmic compression, yielding $L_{\text{agg}}(\mathbf{x}, t) \doteq \log(I_{\text{agg}}(\mathbf{x}, t))$. Accordingly, the event triggering condition is reformulated based on the temporal variation of this spatially filtered response. An event $e_k = (\mathbf{x}_k, t_k, p_k)$ is asynchronously generated whenever the change in $L_{\text{agg}}$ exceeds the contrast threshold $C$:
\begin{equation}
    \Delta L_{\text{agg}}(\mathbf{x}_k, t_k) \doteq L_{\text{agg}}(\mathbf{x}_k, t_k) - L_{\text{agg}}(\mathbf{x}_k, t_{\text{last}}) = p_k C,
    \label{eq:fs_dvs_trigger}
\end{equation}
where $t_{\text{last}}$ is the timestamp of the last triggered event at location $\mathbf{x}_k$. This design ensures that event generation is driven by the spatio-temporal dynamics of a local neighborhood rather than an isolated point, inherently equipping the sensor with frequency-selective capabilities.

While the center-surround structure of biological RGCs provides a robust conceptual blueprint, manually designing a fixed spatial kernel (e.g., a hand-crafted Difference-of-Gaussians filter) is suboptimal for artificial visual perception. Therefore, we parameterize $\mathbf{W}$ as a learnable spatial prior rather than a rigid physical constraint. To enable the task-driven optimization of this frequency-selective sensing frontend, the event generation mechanism must be integrated into a fully differentiable simulation framework. This allows the spatial filter $\mathbf{W}$ to be jointly optimized via end-to-end gradient backpropagation, a process detailed in the following section.

\subsection{Differentiable Event Simulation Framework}
To enable the end-to-end optimization of $\mathbf{W}$ driven by downstream tasks, we must integrate the discrete event generation process into a differentiable framework. 
While Eq. \eqref{eq:fs_dvs_trigger} provides a mathematically elegant constant-contrast model, real-world physical DVS sensors exhibit highly complex non-ideal behaviors. To ensure a high-fidelity sim-to-real transfer, we build our framework upon V2CE~\cite{zhang2024v2ce}, a state-of-the-art neural-network-based event simulator. Unlike purely analytical simulators, V2CE utilizes a data-driven neural architecture to accurately map intensity variations to event streams, yielding simulated outputs that closely match real sensor distributions.

The overall simulation and training pipeline is illustrated in Fig.~1(a). In the forward pass, the input video frames are first processed by the single spatial filter $\mathbf{W}$ to perform spatial aggregation. The filtered frames are then fed into the event simulator to generate an event volume, which is subsequently consumed by the downstream task networks. The original V2CE pipeline contains several discrete and non-differentiable operations, such as explicit voxel detachment and discrete event sampling, which interrupt gradient propagation. To address this, we reformulate the simulation process as a tensor-based differentiable module, in which voxel merging is implemented with in-graph tensor operations, while discrete event-stream generation is replaced by direct aggregation of predicted event voxels into dense event-frame representations. During backpropagation, the modified V2CE simulator acts as a frozen differentiable mapping function. Task-specific loss gradients seamlessly penetrate the frozen simulator and update the weights of the spatial filter $\mathbf{W}$. To ensure training stability, the filter $\mathbf{W}$ is rigorously initialized close to a spatial Dirac delta function (i.e., a value of 1 at the center pixel and small random noise elsewhere), making the FS-DVS equivalent to a conventional single-pixel DVS at the start of training. 

\subsection{Sensor-in-the-Loop Physical Validation}
While the differentiable simulation framework (Sec. 3.3) enables end-to-end optimization, pure digital simulations cannot perfectly replicate the complex non-ideal behaviors of actual physical sensors. To validate that the learned frequency-selective prior $\mathbf{W}$ maintains its noise-resilient properties in the physical world, we construct a sensor-in-the-loop emulation system, as shown in Fig.~1(b).

This system serves as a sensor-in-the-loop proxy for the proposed FS-DVS paradigm. To avoid temporal aliasing, the source videos are first temporally upsampled to a high frame rate. We then establish two parallel physical capture pipelines to conduct a comparative study. In the baseline pipeline (representing conventional DVS), the high-speed original video is directly displayed on a high-refresh-rate monitor and recaptured by a physical dynamic vision sensor. In the FS-DVS pipeline, the video is first convolved with the learned spatial filter $\mathbf{W}$ before being displayed and captured under identical physical sensing conditions.

The critical scientific value of this recapturing process lies in the natural injection of real-world hardware noise and other physical limitations into the event stream. By evaluating these physically captured event streams on identical downstream tasks, we can quantitatively assess the sim-to-real transferability of our paradigm. Ultimately, this sensor-in-the-loop evaluation provides empirical evidence that the pre-trigger spatial aggregation mechanism is a robust and physically viable blueprint for next-generation neuromorphic sensors.

\section{Experiments}

In this section, we systematically evaluate the FS-DVS paradigm through a two-pronged framework: downstream performance validation and in-depth mechanism analysis. First, we demonstrate the efficacy and universality of FS-DVS across three distinct visual tasks spanning global to pixel-level perception. Specifically, we evaluate action recognition and object detection, respectively. Furthermore, we introduce a zero-shot transfer experiment on semantic segmentation, proving the task-agnostic generalization of the learned physical prior without any retraining.

Second, we delve into the underlying physical principles of the proposed paradigm. Through a kernel frequency analysis, we reveal that the learned spatial filters spontaneously align with the mid-frequency enhancement characteristic of the human CSF. Finally, a bandwidth-constrained comparison contrasts FS-DVS against naive hardware threshold adjustments. This confirms that our method's performance gains originate from extracting high-quality, task-relevant information rather than merely inflating the event generation rate.

\subsection{Implementation Details and Evaluation Environments}

\subsubsection{Event Stream Generation and Bandwidth Control}

To validate the proposed paradigm, all event streams used in our experiments are generated through two distinct tracks: \textit{Differentiable Simulation} and \textit{Physical Sensor-in-the-Loop Validation}. In the simulation track, the V2CE simulator is employed to systematically convert consecutive RGB frames into event streams. In the physical validation track, the source videos are first temporally interpolated to 400\,Hz using RIFE~\cite{huang2022realtimeintermediateflowestimation} to approximate continuous physical dynamics. The upsampled videos are then displayed on an LG 27G850A 480\,Hz monitor, and a Prophesee EVK4 HD camera ($1280 \times 720$ resolution) is utilized to physically recapture the scenes. During the sensor-in-the-loop validation, we explicitly control the event bandwidth (i.e., the generated event count) to enable a systematic evaluation of the proposed paradigm. Specifically, the event rate is adjusted by tuning the hardware ON and OFF contrast threshold biases of the Prophesee EVK4 camera.

\subsubsection{Object Detection Setup}

We evaluate object detection on the DSEC-Detection dataset~\cite{9387069}, which consists of 33 training sequences and 8 test sequences. The RGB frames are captured at an original resolution of $1440 \times 1080$ at 20 FPS and are resized to $640 \times 480$ for spatial alignment with the event sensor. For performance reporting, we focus on five representative categories: person, car, bus, truck and two-wheeler.

For both evaluation tracks, the RGB sequences serve as the primary source of intensity signals for generating event streams. In the differentiable simulation track, the V2CE simulator processes 17 consecutive RGB frames to synthesize a spatio-temporal event stream. For the downstream object detection task, this stream is subsequently accumulated into a structured tensor of shape $16 \times 2 \times 480 \times 640$, which serves as the input representation for the RT-DETR detector. In the sensor-in-the-loop validation track, these source RGB sequences are displayed on the monitor and physically recaptured by the Prophesee EVK4 camera. We specifically bypass the native event streams provided by the DSEC dataset because the FS-DVS paradigm requires the spatial filter to be applied strictly prior to the event triggering mechanism. Moreover, deriving all event data from the same source RGB sequences ensures that both the baseline DVS and the proposed FS-DVS share an identical event generation pipeline, thereby guaranteeing a strictly fair comparison.

The downstream RT-DETR~\cite{Zhao_2024_CVPR} detector is trained on a single NVIDIA A800 GPU for 50 epochs using the AdamW optimizer with a batch size of 16. Crucially, prior to optimization, the spatial filter $\mathbf{W}$ is initialized as a spatial Dirac delta function augmented with small random noise (i.e., a value of 1 at the center and near-zero random values elsewhere). This subtle randomness is essential to break weight symmetry during gradient descent, ensuring stable optimization while maintaining a bias-free starting point equivalent to a conventional single-pixel DVS. The training then proceeds via a two-stage strategy. Initially, the detector is trained with the frozen filter to establish a stable baseline representation. Subsequently, a joint end-to-end optimization is applied, allowing the task-driven loss to dynamically refine the frequency-selective evolution of $\mathbf{W}$. During this joint phase, the learning rates are decoupled: $5 \times 10^{-4}$ for the filter, $1 \times 10^{-6}$ for the backbone, and $1 \times 10^{-5}$ for the detection head.

\subsubsection{Action Recognition Setup}
We evaluate action recognition on the UCF101 dataset~\cite{soomro2012ucf101dataset101human}, a widely adopted benchmark comprising 101 categories of daily activities. The source RGB videos are captured at a spatial resolution of $320 \times 240$ with a frame rate of 25 FPS. We randomly sample short temporal clips during both training and evaluation to assess the model's short-term spatio-temporal recognition capabilities.

In the differentiable simulation track, the V2CE simulator processes 17 consecutive RGB frames to synthesize a spatio-temporal event stream, representing a 0.64\,s effective temporal window. For the downstream recognition task, this stream is subsequently accumulated into a structured tensor of shape $16 \times 2 \times 240 \times 320$, which serves as the input representation for the MViT~\cite{Li_2022_CVPR} classifier. In the sensor-in-the-loop validation track, the temporally interpolated source videos are displayed on a high-refresh-rate monitor and physically recaptured by the Prophesee EVK4 camera to obtain the corresponding hardware-generated event streams. 

Following the same hardware (NVIDIA A800 GPU), optimizer (AdamW), and filter initialization protocols as the detection task, the MViT classifier is optimized entirely end-to-end for 50 epochs with a batch size of 12 without a frozen-filter stage. The learning rates are decoupled: $5 \times 10^{-4}$ for the spatial filter, $5 \times 10^{-5}$ for the MViT backbone, and $1 \times 10^{-4}$ for the classification head.

\begin{figure*}[t]
    \centering
    \includegraphics[width=\textwidth]{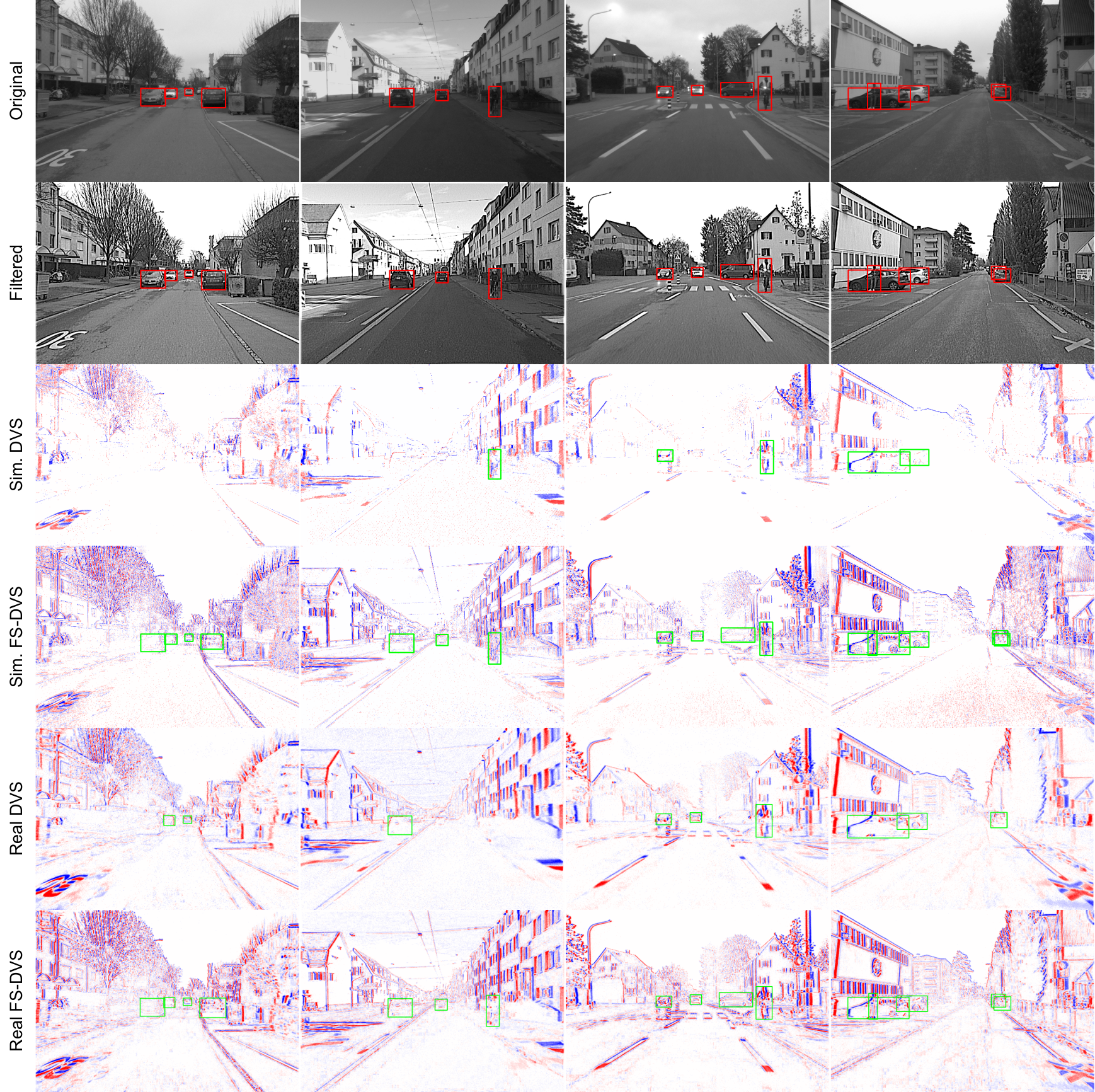}
    \caption{Qualitative comparison of image-domain inputs and event representations with and without spatial filtering. From top to bottom, the rows show the original image, the image processed by the learned $7\times 7$ spatial filter, simulated DVS events, simulated FS-DVS events, real DVS events, and real FS-DVS events. Red and green bounding boxes denote the ground truth (GT) and detector predictions, respectively.}
    \label{fig:qualitative_comparison_detection}
\end{figure*}

\begin{figure}[t]
    \centering
    \includegraphics[width=\columnwidth]{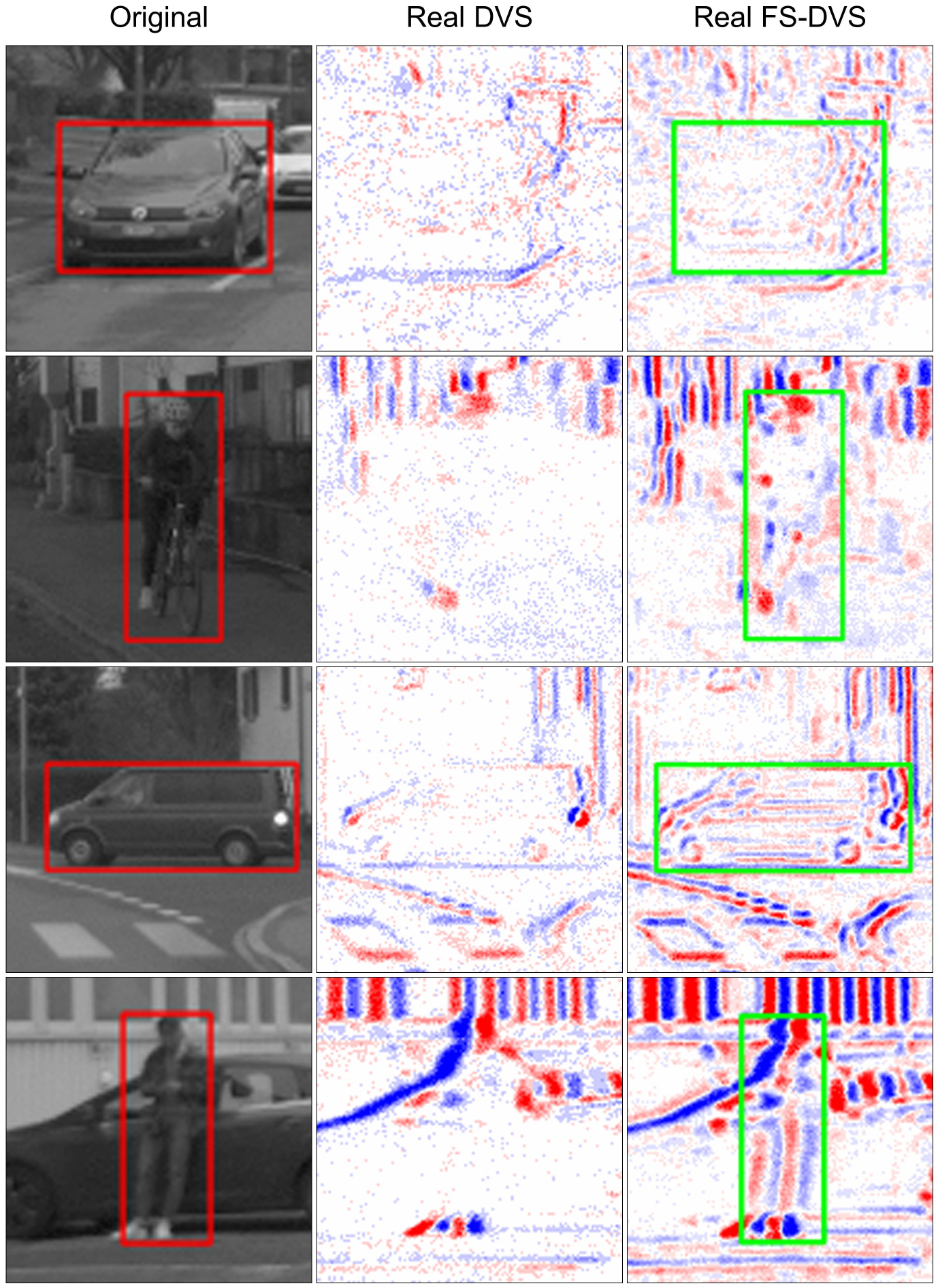}
    \caption{Zoomed-in comparison of real DVS and real FS-DVS event representations in selected challenging object regions. Red and green bounding boxes denote the ground truth (GT) and detector predictions, respectively.}
    \label{fig:real_event_zoom_comparison}
\end{figure}

\subsection{Evaluation on Object Detection}

Quantitative performance is evaluated using the standard COCO mean average precision (mAP) metric~\cite{lin2014microsoft}. As detailed in Table \ref{tab:filter_detection}, the proposed FS-DVS paradigm exhibits significant and consistent superiority over the conventional single-pixel DVS baseline across both simulated and physical environments. In the differentiable simulation track, detection efficacy scales positively with the spatial filter size, peaking at the $7\times7$ configuration with a remarkable +12.3 mAP improvement. Expanding the filter to $9\times9$ yields diminishing returns, indicating a saturation point in spatial context aggregation. Crucially, this algorithmic advantage translates seamlessly to the physical domain. Under sensor-in-the-loop validation, the optimal $7\times7$ FS-DVS maintains a striking +10.8 mAP margin over the hardware baseline, definitively proving the robustness of the learned frequency-selective prior against real-world sensor noise.

Qualitative visualizations in Figs.~\ref{fig:qualitative_comparison_detection} and \ref{fig:real_event_zoom_comparison} explicitly elucidate the underlying physical mechanism driving these quantitative gains. As observed in the conventional DVS rows of Fig.~\ref{fig:qualitative_comparison_detection}, standard event generation relies predominantly upon absolute temporal intensity variations. Consequently, kinematically subtle targets, including distant vehicles and slowly moving pedestrians, frequently fail to exceed the activation threshold. This limitation results in severely sparse and fragmented event responses, leading to subsequently missed detection bounding boxes. In contrast, the proposed FS-DVS paradigm introduces frequency-selective spatial filtering strictly prior to event triggering. As depicted in the second row of Fig.~\ref{fig:qualitative_comparison_detection}, this pre-trigger operation explicitly amplifies task-relevant spatial gradients, which correspond to critical structural contours, without amplifying high-frequency sensor noise and homogeneous background variations.

The zoomed-in physical captures in Fig.~\ref{fig:real_event_zoom_comparison} provide granular empirical evidence for this selective enhancement. In highly challenging scenarios involving the frontal profile of a low-contrast vehicle and a nearly static pedestrian standing beside a car, the physical FS-DVS configuration successfully preserves continuous and highly discriminative object boundaries. In these identical regions, the conventional physical DVS yields negligible valid event activations due to insufficient temporal contrast. By incorporating spatial filtering prior to event generation, FS-DVS significantly enhances the structural completeness and geometric continuity of the resulting event representation. This structurally enriched signal provides the essential topological cues for accurate object localization.

\begin{table}[t]
\centering
\caption{Object Detection Performance with Varying Spatial Filters}
\label{tab:filter_detection}
\setlength{\tabcolsep}{10pt}
\renewcommand{\arraystretch}{1.15}
\begin{tabular}{llcccc}
\toprule
Data & Filter setting & mAP & AP$_{50}$ & AP$_{75}$ & $\Delta$ mAP \\
\midrule
Sim. & w/o & 30.1 & 46.5 & 31.8 & 0.0 \\
Sim. & $3\times3$ & 38.2 & 56.1 & 40.2 & +8.1 \\
Sim. & $5\times5$ & 40.7 & 57.0 & 44.2 & +10.6 \\
Sim. & $7\times7$ & \textbf{42.4} & \textbf{59.8} & \textbf{45.9} & \textbf{+12.3} \\
Sim. & $9\times9$ & 40.3 & 57.5 & 43.5 & +10.2 \\
\midrule
Real & w/o & 32.9 & 48.6 & 35.2 & 0.0 \\
Real & $7\times7$ & \textbf{43.7} & \textbf{63.3} & \textbf{47.1} & \textbf{+10.8} \\
\bottomrule
\end{tabular}\vspace{0.15cm}

\par\parbox{0.9\columnwidth}{\footnotesize \textit{Note}: The w/o setting denotes the conventional DVS baseline without spatial filtering. $\Delta$ mAP indicates the performance gain relative to the baseline within the respective simulated or real-world data track.}
\end{table}

\subsection{Evaluation on Action Recognition}

\begin{table}[t]
\centering
\caption{Action Recognition Performance with Varying Spatial Filters}
\label{tab:filter_action}
\setlength{\tabcolsep}{10pt}
\renewcommand{\arraystretch}{1.15}
\begin{tabular}{llcc}
\toprule
Data & Filter setting & Top-1 Acc. & $\Delta$ Acc. \\
\midrule
Sim. & w/o & 84.09 & 0.00 \\
Sim. & $3\times3$ & 86.59 & +2.50 \\
Sim. & $5\times5$ & 89.01 & +4.92 \\
Sim. & $7\times7$ & \textbf{92.95} & \textbf{+8.86} \\
Sim. & $9\times9$ & 88.70 & +4.61 \\
\midrule
Real & w/o & 87.09 & 0.00 \\
Real & $7\times7$ & \textbf{93.51} & \textbf{+6.42} \\
\bottomrule
\end{tabular}\vspace{0.15cm}

\par\parbox{0.9\columnwidth}{\footnotesize \textit{Note}: The w/o setting denotes the conventional DVS baseline without spatial filtering. $\Delta$ Acc. indicates the Top-1 accuracy gain relative to the baseline within the respective simulated or real-world data track.}
\end{table}

Quantitative evaluation of action recognition is conducted using the Top-1 accuracy metric. As reported in Table \ref{tab:filter_action}, FS-DVS demonstrates a decisive advantage over the conventional DVS baseline in overall classification performance. Consistent with the object detection results, recognition accuracy exhibits a positive correlation with the spatial filter size up to the optimal $7\times7$ configuration, which yields a substantial +8.86\% improvement in the simulation track. Beyond this scale, the performance marginally degrades, reaffirming the saturation point in spatial context aggregation. Crucially, this algorithmic superiority successfully transfers to the physical sensor-in-the-loop validation track. The FS-DVS configuration achieves a significant +6.42\% accuracy gain over the hardware baseline, definitively proving its efficacy and robustness in real-world scenarios.

\begin{figure*}[htbp]
    \centering
    \includegraphics[width=\textwidth]{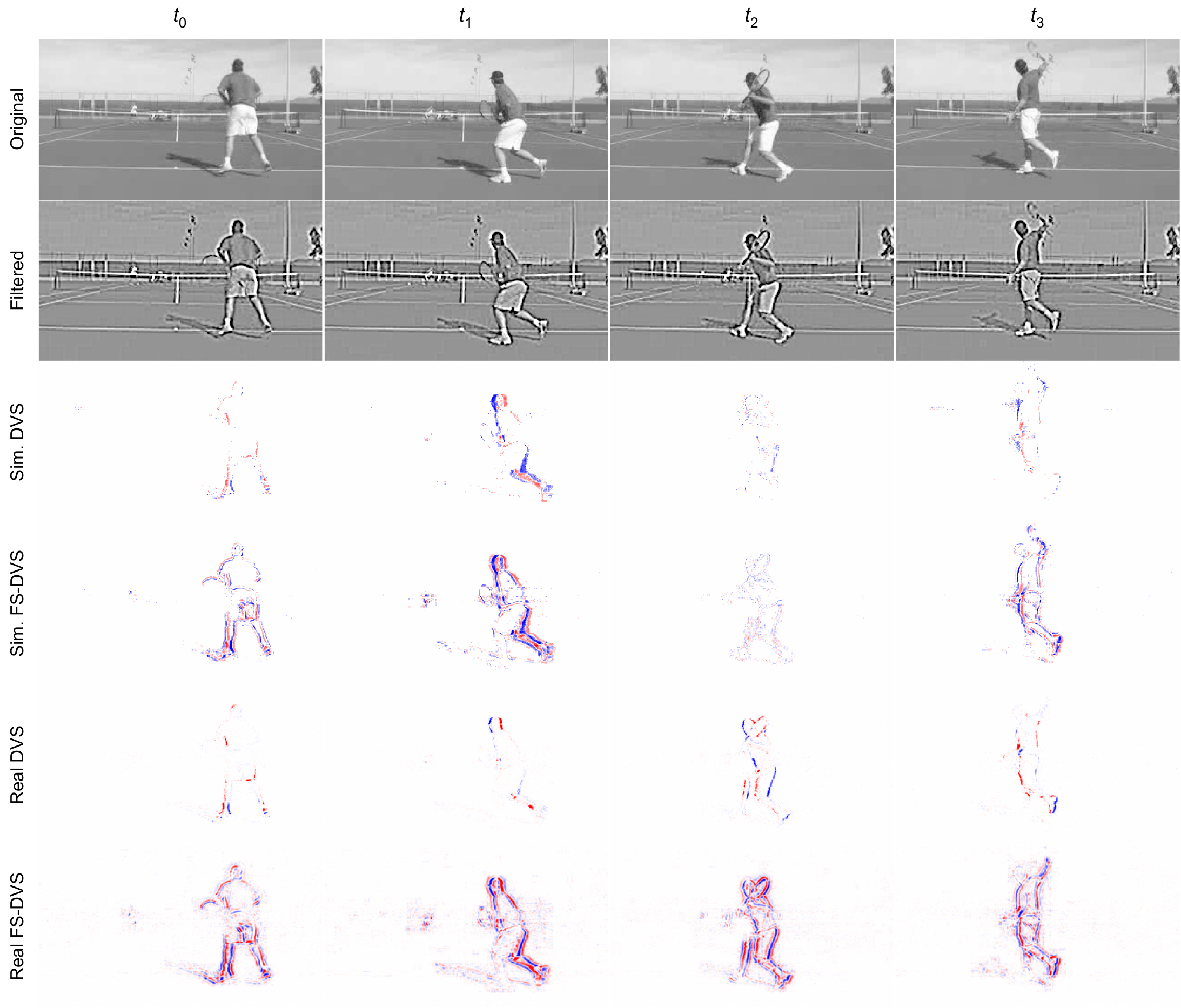}
    \caption{Visualization of representative frames for event-based action recognition. Each column shows a representative time instant within the same action sequence. From top to bottom, the rows present the original image, the filtered image processed by the learned $7\times7$ spatial filter, simulated DVS events, simulated FS-DVS events, real DVS events, and real FS-DVS events.}
    \label{fig:qualitative_comparison_action}
\end{figure*}

The qualitative visualizations in Fig.~\ref{fig:qualitative_comparison_action} explicitly elucidate how this pre-trigger spatial aggregation benefits global motion perception. In the conventional DVS frames, event generation is strictly dominated by regions with high-velocity relative motion (e.g., the rapidly swinging tennis racket), while slower-moving body parts fail to trigger sufficient events. This discrepancy results in temporally fragmented and spatially incomplete kinematic trajectories. By explicitly amplifying task-relevant spatial gradients prior to event generation, FS-DVS successfully preserves complete human contours and continuous motion trajectories. This structural and geometric completeness ensures that the downstream classifier receives dense, fine-grained kinematic cues. Ultimately, these results confirm that the benefits of the learned spatial prior extend far beyond local spatial localization, fundamentally enhancing the representation of global spatio-temporal patterns.

\subsection{Transferring the Learned Spatial Filter to Dense Semantic Segmentation}

To evaluate whether the learned spatial filter can generalize across different visual tasks, we investigate its cross-task zero-shot transfer capability on dense pixel-level semantic segmentation using the DSEC-Semantic benchmark. Unlike previous experiments, we conduct evaluations directly on the physical sensor-in-the-loop track to ensure the most authentic assessment of cross-task robustness. Specifically, the optimal $7\times7$ spatial filter learned exclusively from the object detection task is directly frozen and deployed as a pre-trigger filter. The physically captured event streams are subsequently processed by a standard Mask2Former network~\cite{Cheng_2022_CVPR}.

\begin{figure}[t]
    \centering
    \includegraphics[width=\columnwidth]{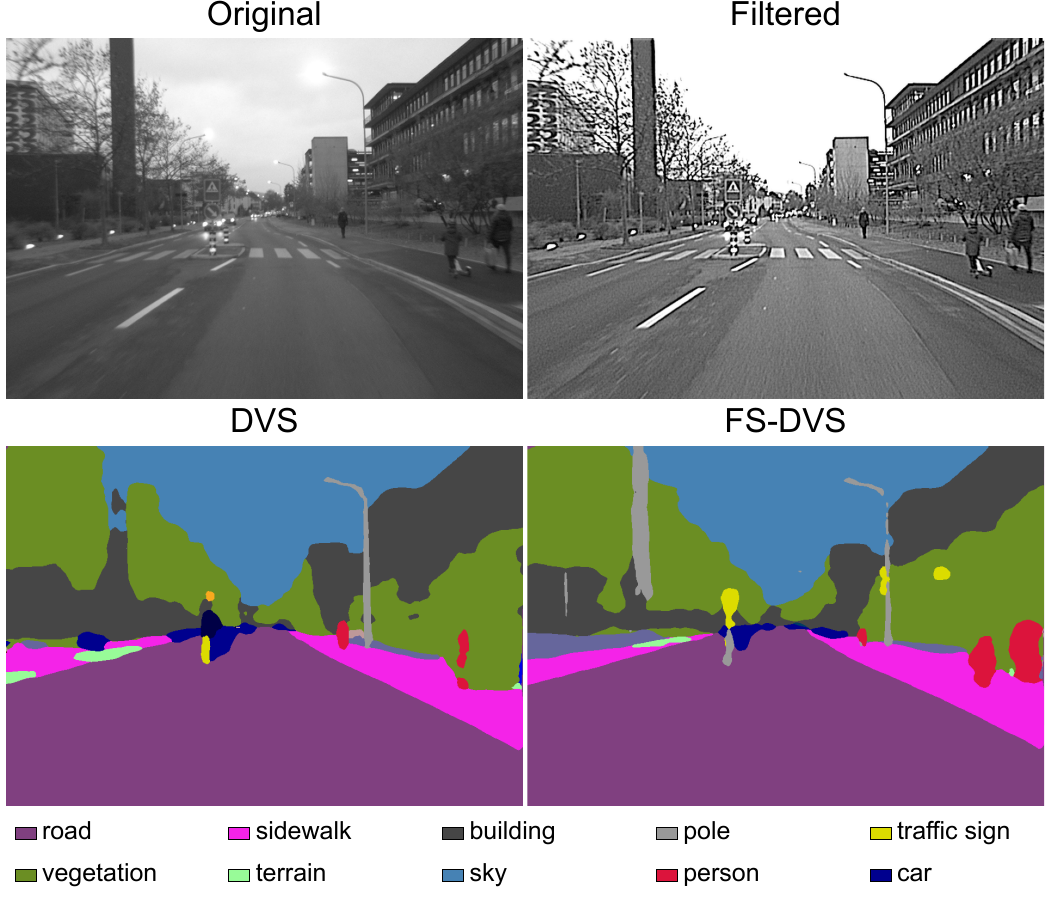}
    \caption{Comparison of zero-shot semantic segmentation results. The top row shows the visual representations before and after spatial filtering, while the bottom row shows the corresponding segmentation maps.}
    \label{fig:segmentation_visualization}
\end{figure}

Quantitative results on real-world event data are reported in Table~ \ref{tab:filter_segmentation}. The zero-shot application of the detection-derived spatial filter yields a substantial +4.77\% improvement in the mean Intersection over Union (mIoU) metric compared to the conventional physical DVS baseline. 
Qualitative results in Fig.~\ref{fig:segmentation_visualization} further illustrate these improvements. The pre-trigger spatial filtering significantly enhances the structural continuity of object boundaries, particularly for challenging instances such as distant pedestrians and slender traffic signs. Importantly, achieving these substantial gains without task-specific retraining implies that the learned spatial filtering operates as a general low-level geometric feature extractor, rather than a pattern explicitly overfitted to detection tasks.

\begin{table}[t]
\centering
\caption{Semantic Segmentation Performance via Zero-Shot Spatial Filter Transfer}
\label{tab:filter_segmentation}
\setlength{\tabcolsep}{10pt}
\renewcommand{\arraystretch}{1.15}
\begin{tabular}{lcccc}
\toprule
Filter setting & mIoU & mAcc & PA & $\Delta$ mIoU \\
\midrule
w/o & 39.30 & 47.90 & 87.09 & 0.00 \\
$7\times7$ & \textbf{44.07} & \textbf{53.08} & \textbf{89.27} & \textbf{+4.77} \\
\bottomrule
\end{tabular}\vspace{0.15cm}

\par\parbox{0.9\columnwidth}{\footnotesize \textit{Note}: The w/o setting denotes the conventional DVS baseline under physical capture. The $7\times7$ spatial filter is directly transferred from the object detection task without any task-specific retraining. $\Delta$ mIoU indicates the performance gain relative to the baseline.}
\end{table}

\subsection{Evolutionary Convergence: The Learned Filter and Human Vision}

\begin{figure}[t]
    \centering
    \includegraphics[width=\columnwidth]{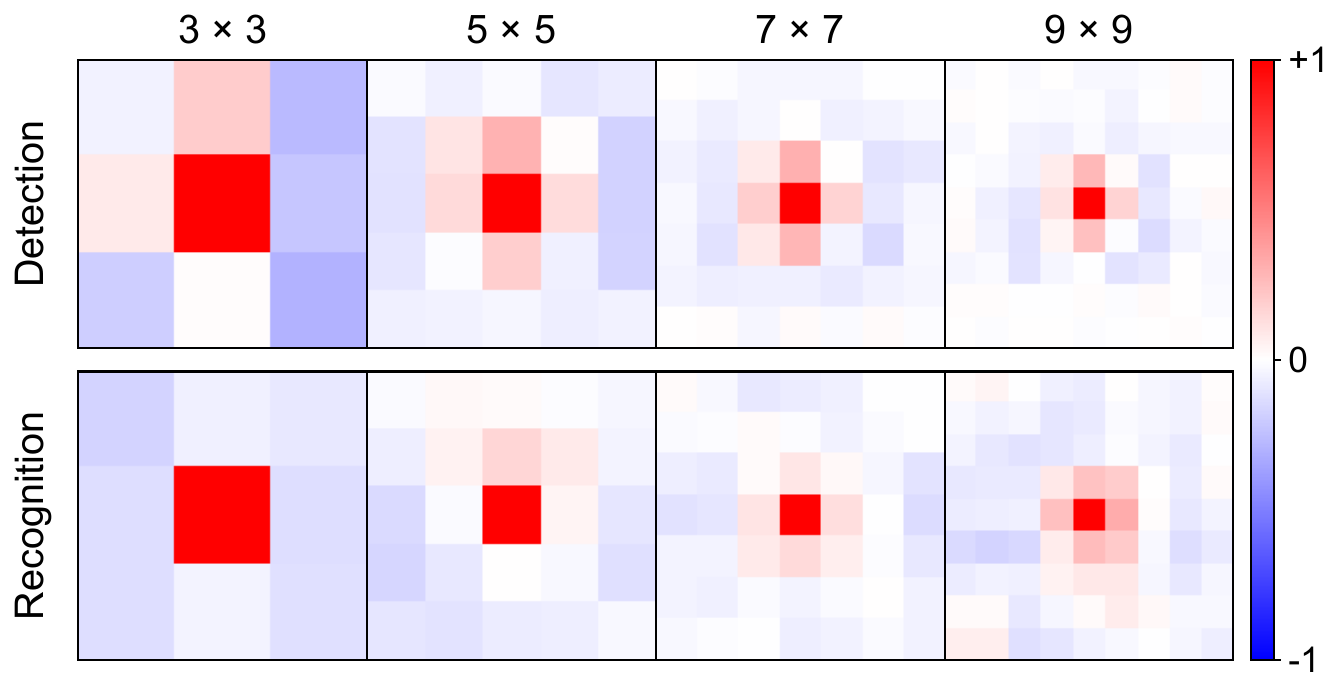}
    \caption{Learned spatial filters exhibit an antagonistic center-surround organization that is functionally analogous to the receptive fields of retinal ganglion cells (RGCs).}
    \label{fig:kernel_weights}
\end{figure}

The zero-shot transferability suggests that the learned spatial filter captures universal visual primitives rather than task-specific artifacts. As shown in Fig.~\ref{fig:kernel_weights}, despite the distinct nature of the two tasks, the optimized filters consistently converge toward an antagonistic center-surround topology. This emerging structure mirrors the biological receptive fields of ON-center/OFF-surround retinal ganglion cells (RGCs), which utilize lateral inhibition to maximize local contrast and extract salient edge information—the most stable structural primitives across diverse natural scenes.

\begin{figure}[t]
    \centering
    \includegraphics[width=\columnwidth]{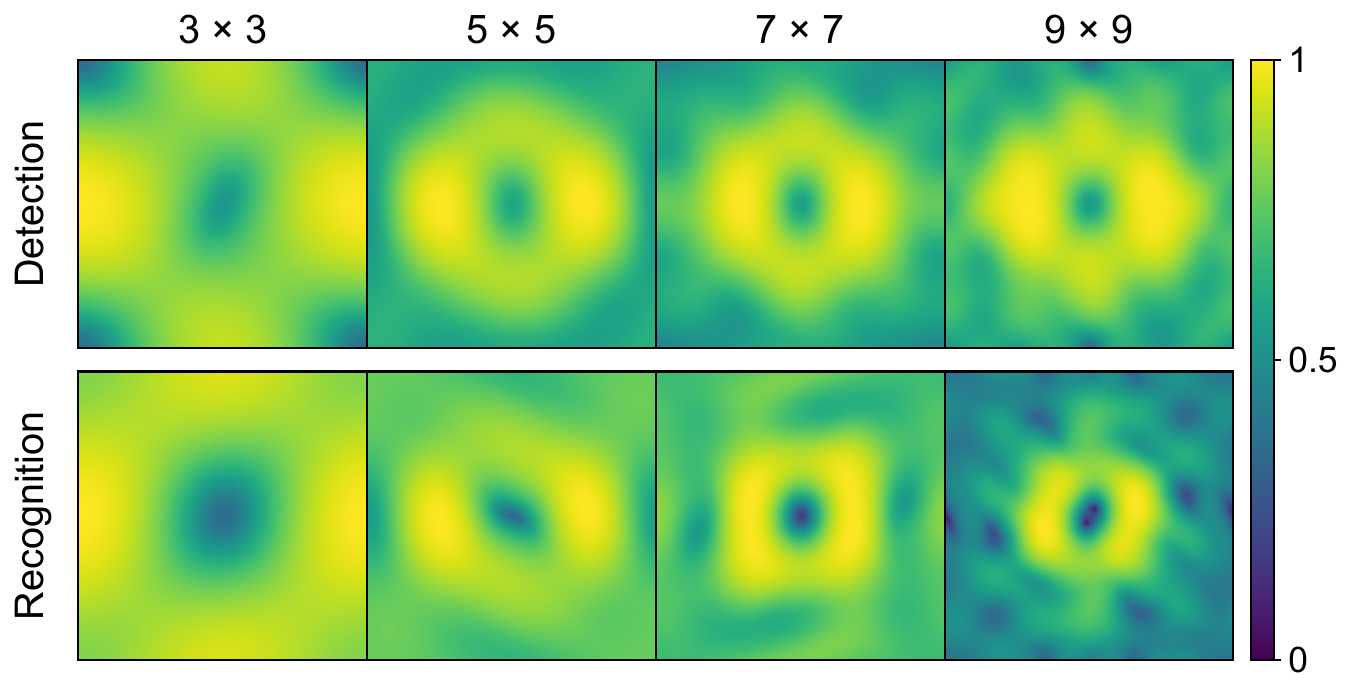}
    \caption{Two-dimensional FFT magnitude spectra of the learned spatial filters, showing consistent ring-like energy distributions across different tasks.}
    \label{fig:fft_spectra}
\end{figure}

\begin{figure}[t]
    \centering
    \includegraphics[width=\columnwidth]{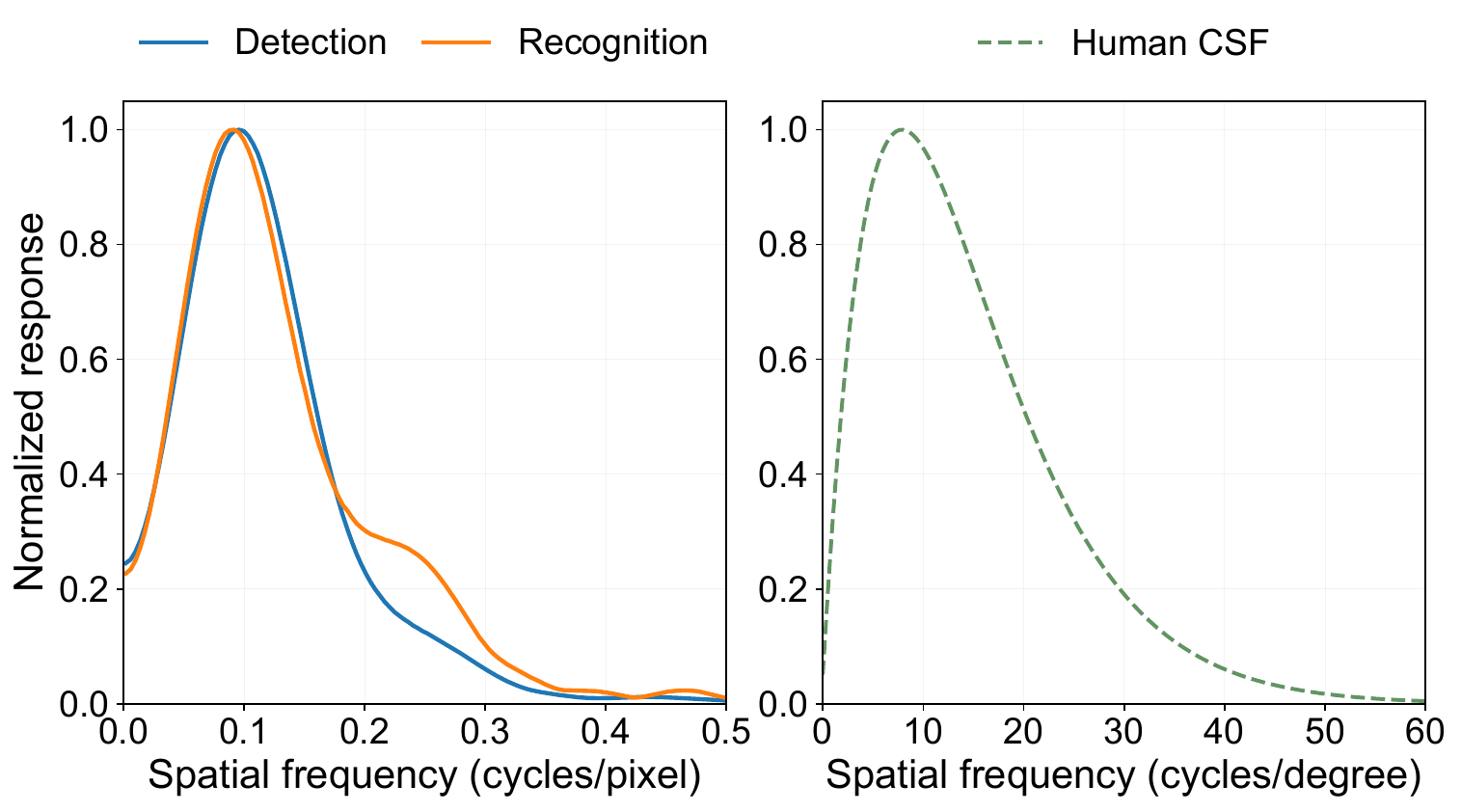}
    \caption{
Radial frequency responses of the learned spatial filters and a normalized Mannos--Sakrison contrast sensitivity function (CSF) reference.}
    \label{fig:frequency_response}
\end{figure}

To quantify this sensing strategy, we compute the two-dimensional Fast Fourier Transform (FFT) magnitude spectra of the learned filters (Fig.~\ref{fig:fft_spectra}). The spectra consistently exhibit ring-shaped energy distributions, indicating band-pass responses concentrated in the intermediate-frequency range. We further derive radial frequency responses from the 2D spectra and compare them with a normalized Mannos--Sakrison contrast sensitivity function (CSF)~\cite{1055250}, which serves as a representative reference for human visual sensitivity. Since the filter responses and the CSF are defined in different physical units, we focus on comparing their relative shapes. Specifically, the response magnitude of each curve is normalized to its own maximum, while the frequency axes are preserved in their native units. As illustrated in Fig.~\ref{fig:frequency_response}, the learned filters spontaneously ``rediscover'' the biological band-pass trend: they aggressively suppress low-frequency illumination redundancy and high-frequency stochastic noise to focus exclusively on the informative intermediate-frequency band. Notably, without imposing any explicit frequency-domain constraint or biological prior, the radial responses learned from detection and action recognition also exhibit a high degree of overlap, with their peak responses located at nearly identical spatial frequencies. This evolutionary convergence suggests that task-driven AI optimization independently arrives at the same sensing principles optimized by millions of years of biological evolution. Such biological plausibility provides a rigorous explanation for the robust generalization observed earlier: by converging toward a CSF-like band-pass prior, FS-DVS captures the core geometric features essential for resilient perception across diverse visual tasks.

\subsection{Bandwidth Efficiency and Information-Theoretic Analysis}
\begin{figure}[t]
    \centering
    \includegraphics[width=\columnwidth]{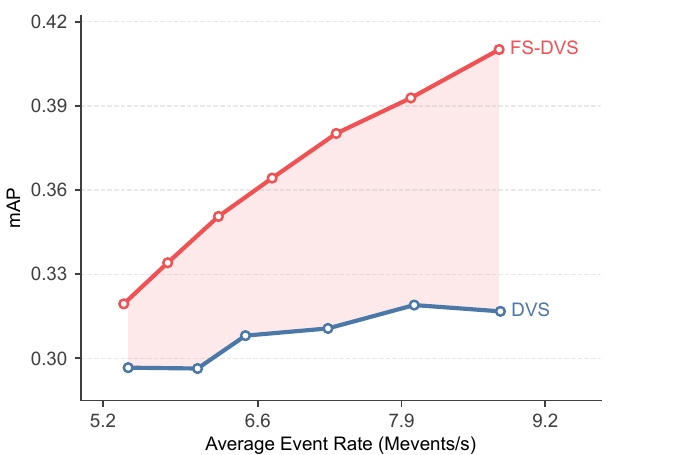}
    \caption{Detection mAP versus average event rate for DVS and FS-DVS.}
    \label{fig:bandwidth_evaluation}
\end{figure}

To verify that the gains of FS-DVS stem from optimized filtering rather than a trivial increase in sensitivity, we evaluate detection performance under matched event rates (Fig.~\ref{fig:bandwidth_evaluation}). While FS-DVS exhibits a steady performance climb with increasing bandwidth, the conventional DVS suffers a sharp accuracy collapse at high sensitivity. This ``noise avalanche'' confirms that blindly lowering thresholds in standard DVS only accumulates stochastic noise; in contrast, the learned spatial prior in FS-DVS ensures that additional bandwidth is utilized for structurally informative cues.

Beyond task-specific empirical performance, we provide an information-theoretic evaluation of reconstruction fidelity by computing the mutual information (MI) between the ground truth and the frames reconstructed via E2VID~\cite{10462903}. FS-DVS consistently outperforms conventional DVS, yielding an average relative MI gain of 8.16\% (increasing from 0.98 to 1.06 bits per pixel). This consistent improvement explicitly demonstrates that the learned spatial filter preserves richer, more recoverable structural and geometric information, fundamentally enhancing the information density of the event stream.

\subsection{Hardware Feasibility Discussion}

While this work validates the frequency-selective spatial priors through a physical sensor-in-the-loop setup, directly embedding these learned filters into the dynamic vision sensor represents a promising future direction. Existing neuromorphic designs, such as the Center-Surround DVS (CS-DVS)~\cite{delbruck2022utility}, have demonstrated that hardware-level spatial smoothing can be efficiently achieved using 2D resistive meshes. However, the basic CS-DVS only extracts high-frequency information by subtracting a broad smoothed background directly from a strict single-pixel center. 

\begin{figure}[t]
    \centering
    \includegraphics[width=\columnwidth]{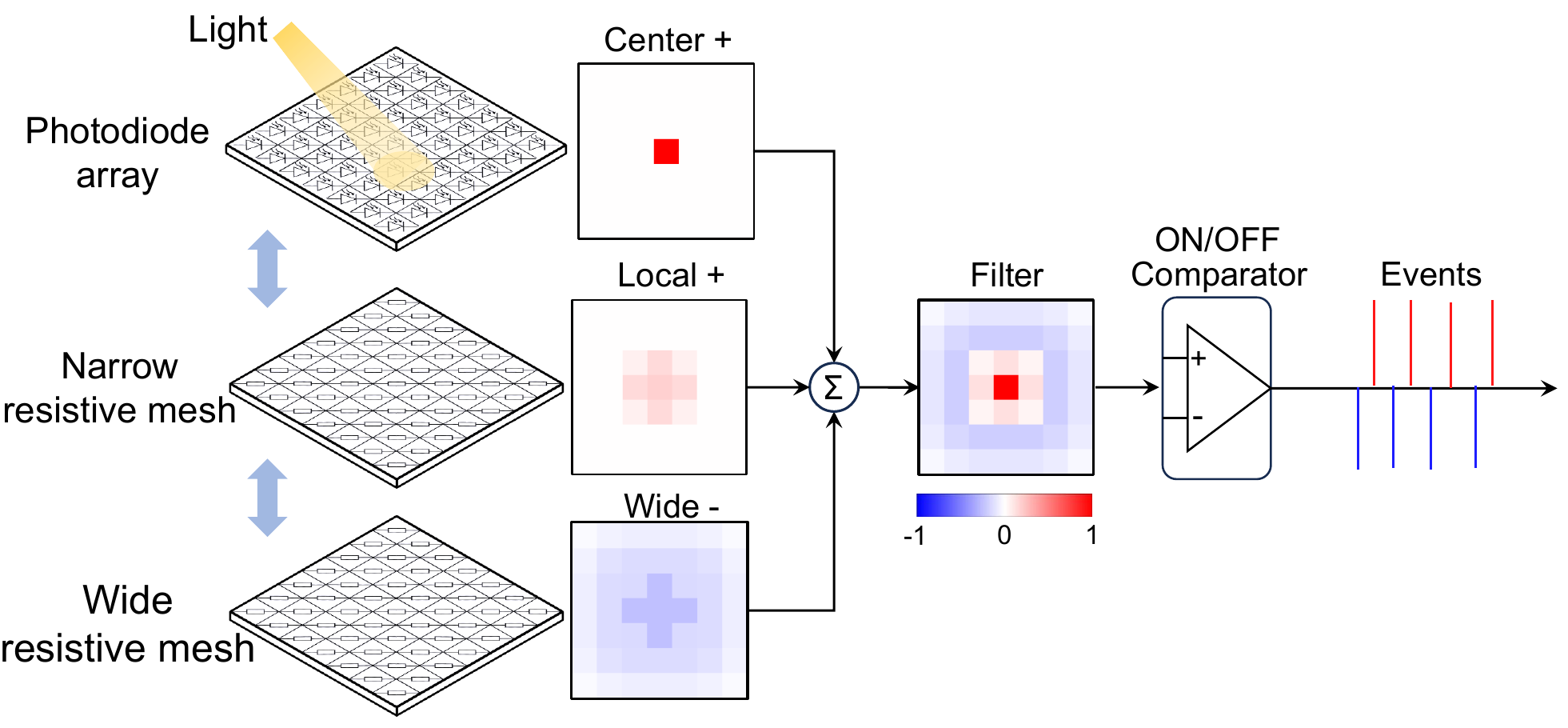}
    \caption{Conceptual dual-mesh implementation of FS-DVS. The narrow and wide resistive meshes respectively realize local positive aggregation and surround suppression, approximating the learned band-pass kernel before event triggering.}
    \label{fig:hardware_feasibility}
\end{figure}

To physically synthesize our learned mid-frequency enhancement kernels, we hypothesize that this hardware paradigm could potentially be extended to a dual-resistive-mesh architecture (Fig.~\ref{fig:hardware_feasibility}). Specifically, the raw signal from each pixel is concurrently fed into two parallel resistive networks with different diffusion ranges (controlled by lateral resistance and transverse transconductance), denoted as the narrow resistive mesh and the wide resistive mesh. The narrow resistive mesh limits the signal spread to a small neighborhood to form the positive central response (e.g., $3 \times 3$), while the wide resistive mesh allows the signal to spread further to form the broader negative surround (e.g., $7 \times 7$). By simply adding the pixel's own signal to the output of the narrow resistive mesh and subtracting that of the wide resistive mesh, the hardware naturally acts as a spatial band-pass filter. 

Notably, this analog network represents just one potential realization. Since our ablation study demonstrates that optimal performance is achieved at compact kernel sizes (e.g., $5 \times 5$ or $7 \times 7$), this pre-trigger filtering can also be readily implemented via low-overhead ASICs~\cite{9897354} or zero-power computational optics (e.g., metalenses)~\cite{Wu:24}. Ultimately, the efficiency of such a compact mechanism proves that the FS-DVS paradigm offers a highly practical and versatile architectural blueprint for next-generation neuromorphic vision sensors.

\section{Conclusion}
In this work, we introduced FS-DVS, a frequency-selective sensing paradigm that fundamentally resolves the structural incompleteness and noise susceptibility of conventional dynamic vision sensors. By integrating a learnable spatial filter strictly prior to the event generation process and optimizing it through a fully differentiable simulation framework, we transformed event sensing from a rigid, per-pixel independent triggering mechanism into a spatially integrated, data-driven encoding process. A profound discovery of our study is that, driven purely by downstream visual tasks, these learned spatial filters spontaneously evolve into a center-surround topology. This emergence closely aligns with the mid-frequency enhancement of the human contrast sensitivity function, suggesting that biological spatial aggregation constitutes a highly effective strategy for maximizing information acquisition. Furthermore, extending beyond pure simulation, we conducted a sensor-in-the-loop verification where all signals were captured by a physical dynamic vision sensor. By naturally incorporating real-world hardware noise and non-ideal sensor characteristics, this physical evaluation further validates the effectiveness and zero-shot transferability of our learned spatial priors across diverse perception tasks. Ultimately, with a potential in-sensor hardware implementation, the proposed FS-DVS paradigm provides a highly practical and generalizable reference for the future development of neuromorphic vision sensors.



%

\bibliographystyle{IEEEtran}
\bibliography{reference}

\newpage

\vfill

\end{document}